\documentclass[letterpaper, 10 pt, conference]{ieeeconf}  

\IEEEoverridecommandlockouts                              

\let\labelindent\relax

\usepackage[utf8]{inputenc} 
\usepackage[T1]{fontenc}    
\usepackage{url}            
\usepackage{booktabs}       
\usepackage{nicefrac}       
\usepackage{microtype}      
\usepackage{graphics,graphicx,color}
\usepackage{tabularx} 
\usepackage{algorithm}
\usepackage{enumitem}

\usepackage{amsfonts}       
\usepackage{amsmath}       
\usepackage{amssymb}

\usepackage{algpseudocode}

\newcommand{\figref}[1]{Fig.~\ref{#1}}    
\newcommand{\Tabref}[1]{Table~\ref{#1}}

\newcommand{\secref}[1]{Sec.~\ref{#1}} 

\usepackage{xspace}
\makeatletter
\DeclareRobustCommand\onedot{\futurelet\@let@token\@onedot}
\def\@onedot{\ifx\@let@token.\else.\null\fi\xspace}
\makeatother

\usepackage{xr-hyper}
\makeatletter
\newcommand*{\addFileDependency}[1]{
  \typeout{(#1)}
  \@addtofilelist{#1}
  \IfFileExists{#1}{}{\typeout{No file #1.}}
}
\makeatother

\usepackage{xcolor}

\definecolor{ourorange}{HTML}{e19c24}
\definecolor{ourgreen}{HTML}{97b567}
\definecolor{ourred}{HTML}{ec6235}
\definecolor{ourblue}{HTML}{5e81b5}
\definecolor{ourgrey}{HTML}{919191}

\usepackage{subcaption}

\DeclareMathAlphabet{\mathsfit}{\encodingdefault}{\sfdefault}{m}{sl}
\SetMathAlphabet{\mathsfit}{bold}{\encodingdefault}{\sfdefault}{bx}{n}

\makeatletter
\let\NAT@parse\undefined
\makeatother

\usepackage{hyperref}
\usepackage{url}

\title{\LARGE \bf
Symmetry-Guided Memory Augmentation for Efficient Locomotion Learning
}

\author{Anonymous Authors}

\author{Kaixi Bao$^{1}$, Chenhao Li$^{1,2,3}$, Yarden As$^{2,3}$, Andreas Krause$^{3}$ and Marco Hutter$^{1}$
\thanks{$^{1}$Department of Mechanical and Process Engineering, ETH Zurich, Switzerland.}%
\thanks{$^{2}$ETH AI Center, ETH Zurich, Switzerland.}%
\thanks{$^{3}$Department of Computer Science, ETH Zurich, Switzerland.}%
\thanks{Mail: \tt\small \{kaibao, chenhli, yardas, krausea, mahutter\}@ethz.ch}
}

\begin{document}

\maketitle
\thispagestyle{empty}
\pagestyle{empty}

\begin{abstract}
Training reinforcement learning (RL) policies for legged locomotion often requires extensive environment interactions, which are costly and time-consuming.
We propose Symmetry-Guided Memory Augmentation (SGMA), a framework that improves training efficiency by combining structured experience augmentation with memory-based context inference.
Our method leverages robot and task symmetries to generate additional, physically consistent training experiences without requiring extra interactions.
To avoid the pitfalls of naive augmentation, we extend these transformations to the policy's memory states, enabling the agent to retain task-relevant context and adapt its behavior accordingly.
We evaluate the approach on quadruped and humanoid robots in simulation, as well as on a real quadruped platform.
Across diverse locomotion tasks involving joint failures and payload variations, our method achieves efficient policy training while maintaining robust performance, demonstrating a practical route toward data-efficient RL for legged robots.

\end{abstract}

\section{INTRODUCTION}
Reinforcement learning (RL) has shown impressive results in training legged robots to perform agile and adaptive locomotion behaviors~\cite{miki2022perceptiveLoco, lee2020LearnQuadrupedalLoco, kumar2021rma, hoeller2024parkour, li2023learning}.
However, achieving such performance typically requires millions of interactions with the environment~\cite{rudin2022learning, mittal2023orbit}.
While these interactions are usually collected in simulation, prevailing approaches still rely on exposing the agent to a wide range of explicitly randomized task variations—such as altered payloads, terrain properties, or joint failures—so that it can learn robust behaviors~\cite{li2023versatile}.
This strategy is inherently sample-inefficient: the agent wastes interactions collecting data for variations whose outcomes could have been obtained by exploiting the inherent symmetries and other structures of the robot and task.
By not leveraging these inductive biases, much of the learning effort is spent on redundant experiences that could instead be simulated without additional environment interaction~\cite{Mittal2024SymmetryCF}.

To address this challenge, we introduce Symmetry-Guided Memory Augmentation (SGMA), a framework that combines structured experience augmentation with memory-based context inference.
Our method leverages the inherent morphological and task symmetries in legged robots to generate additional training experiences without requiring extra interactions.
Beyond transforming observations and actions, SGMA extends these augmentations to the policy’s hidden memory states, ensuring that the agent can retain task-relevant context rather than collapsing to conservative strategies~\cite{duan2016RL2}.

\begin{figure}[t]
\centering
\includegraphics[width=1.0\columnwidth]{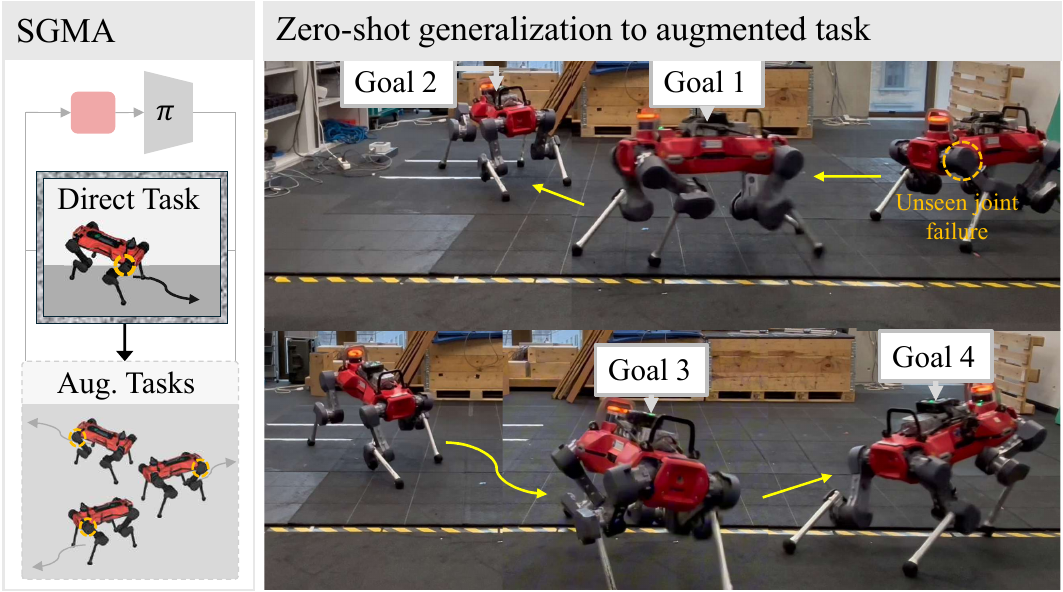}
\caption{\textbf{SGMA} policy in a locomotion task with a right-hind hip abduction adduction (RH HAA) joint failure. By leveraging symmetry-guided memory augmentation, the robot adapts its behavior to track multiple goals under a condition not explicitly encountered during training. More supplementary videos and implementation details are available on our project webpage \url{https://sites.google.com/view/eth-sgma}.}
\label{fig:intro}
\end{figure}

We evaluate SGMA on quadruped and humanoid robots in simulation and on a physical quadruped platform.
Across locomotion tasks involving joint failures and payload variations, SGMA consistently improves training efficiency while preserving robust task performance (\figref{fig:intro}).
Compared to baselines trained with task randomization or naive augmentation, SGMA enables more adaptive locomotion behaviors with fewer environment interactions.

In summary, our contributions include:
\begin{itemize}
    \item We propose SGMA, a principled method that combines symmetry-aware experience augmentation with memory-based context modeling.
    \item We demonstrate that SGMA improves training efficiency by avoiding redundant interactions while preserving adaptability in partially observable settings.
    \item We validate SGMA across both simulation and hardware, showing consistent benefits on quadruped and humanoid locomotion tasks under diverse conditions.
\end{itemize}

\section{RELATED WORK}
\subsection{Leveraging Structures in Legged Locomotion} 

A recurring theme in robot learning is the integration of known physical principles and structural regularities into the learning process~\cite{cai2021physics}.
Such inductive biases reduce the burden of purely data-driven exploration and enable models to capture essential dynamics with fewer interactions.
In legged locomotion, foot-placement dynamics is used to constrain the action space, leading to substantial gains in data efficiency through task-specific mechanics~\cite{yang2020data}.
Parameterized trajectory generators, such as central pattern generators, further reduce exploration from joint space to compact phase–offset representations of predefined body and foot trajectories, forming the basis of several state-of-the-art locomotion controllers~\cite{miki2022perceptiveLoco, zhang2024learning, zhang2025motion}.
Similar ideas using frequency-domain parameterization yield more efficient learning in motion tracking~\cite{li2024fld}, while explicit modeling of granular media interactions improves quadruped locomotion on deformable terrain~\cite{choi2023learning}.
At the dynamics modeling level, combining rigid-body dynamics with differentiable simulation~\cite{song2024learning} or blending analytical formulations with learned residuals in semi-structured Lagrangian models~\cite{levy2024learning} achieves high sample efficiency while preserving physical consistency.

\subsection{Symmetry-Aware Augmentation in RL}

Prior work~\cite{ordonez2024morphological} has studied the inherent morphological symmetries present in many robotic systems, arising from structural features including symmetric mass distributions, duplicated limbs and replicated kinematic chains.
These symmetries introduce structured regularities in both the system dynamics and observation spaces, which have been leveraged in RL via experience augmentation to guide exploration and promote symmetry-invariant behavior in legged locomotion tasks~\cite{Mittal2024SymmetryCF, Su2024LeveragingSym}.
In robotic manipulation, symmetry-aware augmentation has also been applied in conjunction with Hindsight Experience Replay (HER)~\cite{andrychowicz2017HER} to accelerate learning~\cite{lin2020invariantTER}.
Building on this insight, we further explore how symmetry-induced inductive biases can be exploited to improve training efficiency by eliminating the need for redundant environment interactions.

\subsection{Context Inference in RL}

In many robotic tasks, the environment is only partially observable: object properties in manipulation or terrain characteristics in locomotion are often hidden from direct sensing.
One strategy is explicit context inference, where privileged information available during training is distilled into policies that must operate without it at test time.
Such teacher–student frameworks leverage environment parameters or unobservable robot states to guide policies under partial observability, and have been successfully applied in quadrupedal locomotion and dexterous manipulation~\cite{lee2020LearnQuadrupedalLoco, kumar2021rma, qi2023general}.
Complementary to this, implicit inference methods rely on interaction history to capture latent task context.
Early approaches simply concatenated past observations, while more advanced methods employ recurrent networks and memory mechanisms~\cite{duan2016RL2, wang2016learnToRL}, which encode hidden dynamics from trajectories.
These approaches have been used to learn dynamics priors for rapid online adaptation~\cite{nagabandi2018adaptMetaRL} and to develop morphology-agnostic locomotion policies~\cite{zargarbashi2024metaloco}.

\section{PRELIMINARIES}
\subsection{Partially Observable Markov Decision Process}
We formulate the problem as a Partially Observable Markov Decision Process (POMDP), defined by the tuple
\(
\mathcal{M} = \left ( \mathcal{S}, \mathcal{O}, \mathcal{A}, p, r, \gamma, \rho_0 \right ),
\)
where \( \mathcal{S}, \mathcal{O}, \mathcal{A} \) denote the set of states, observations, and actions, respectively.
The transition dynamics are specified by \( p : \mathcal{S} \times \mathcal{A} \times \mathcal{S} \to \mathbb{R}_{\geq 0}\) and reward function by \( r : \mathcal{S} \times \mathcal{A} \to \mathbb{R} \). 
The discount factor is given by \( \gamma \), and the initial state distribution by \( \rho_0 : \mathcal{S} \to \mathbb{R}_{\geq 0} \).
In a POMDP, the agent does not have direct access to the full system state and instead receives partial observations.
In our setting, the latent state contains unobservable environment context, such as unknown payloads or other hidden task parameters, which remain fixed throughout each episode.
The agent’s objective is to learn a policy \( \pi \) that maximizes the expected discounted return:
\[
J(\pi) = \mathbb{E}_{\tau \sim p_\pi(\tau)} \left[ \sum_{t=0}^\infty \gamma^t r(s_t, a_t) \right],
\]
where \( \tau = (s_0, a_0, s_1, \dots) \) represents a trajectory sampled from the POMDP \( \mathcal{M} \) under policy \( \pi \).

\subsection{Symmetric Augmentations}
\label{sec:symmetric_tasks}
We consider a family of tasks that are symmetric at the trajectory level under the action of a group \(\mathbb{G}\).
Formally, let a task $T'$ be associated with the POMDP
\(\mathcal{M}(T') = \left ( \mathcal{S}_{T'}, \mathcal{O}_{T'}, \mathcal{A}_{T'}, p_{T'}, r_{T'}, \gamma_{T'}, \rho_{0,T'} \right ) \), and task $T$ with
\(\mathcal{M}(T) = \left ( \mathcal{S}_{T}, \mathcal{O}_{T}, \mathcal{A}_{T}, p_{T}, r_{T}, \gamma_{T}, \rho_{0,T} \right ) \).
We say that $T'$ is symmetric to $T$ if there exists a transformation \(g \in \mathbb{G}\) acting on states and actions such that
\(g \cdot s \in \mathcal{S}_{T'}, g \cdot a \in \mathcal{A}_{T'}\),
and, for all \(s, s' \in \mathcal{S}_{T}\) and \(a \in \mathcal{A}_{T}\), both the transition dynamics and the reward function remain invariant under $g$:
\[
\begin{aligned}
p_{T'}(g \cdot s' \mid g \cdot s, g \cdot a) &= p_T(s' \mid s, a), \\
r_{T'}(g \cdot s, g \cdot a) &= r_T(s, a), \\
\end{aligned}
\]
For such symmetric tasks, the optimal policy is \(\mathbb{G}\)-equivariant~\cite{wang2022SO2equivariant}, satisfying
\[
\pi_{T'}^*(g \cdot s) = g \cdot \pi_T^*(s), 
\quad \forall g \in \mathbb{G},~ s \in \mathcal{S}_{T}.
\]

Intuitively, many robotic systems naturally exhibit such symmetries.
For example, a quadruped robot with a failed left-front knee joint has a symmetric counterpart where the right-front knee fails, and the corresponding locomotion strategy can be obtained by reflecting the trajectory across the sagittal plane~\cite{Mittal2024SymmetryCF}.
These structured invariances provide a principled means of generating new, physically consistent training experiences without requiring additional environment interactions.


\section{APPROACH}
\subsection{Experience Augmentation via Symmetry}

We exploit the symmetric structure of tasks (\secref{sec:symmetric_tasks}) to improve sample efficiency by dividing each family of tasks into a set of \textit{direct tasks} and their corresponding \textit{augmented tasks}.
During training, the agent interacts only with the direct tasks, while experiences for the augmented tasks are generated in parallel by applying symmetry transformations to the collected trajectories.
This strategy allows the policy to be exposed to a wider variety of task conditions without requiring additional environment interactions.  

Formally, let \( \tau = \{(o_t, a_t, o_{t+1}, r_t)\}_{t=0}^N \) be a trajectory collected by the current policy $\pi$ on a direct task $T$.
For a group element $g = (g_o, g_a) \in \mathbb{G}$, where $g_o : \mathcal{O}_T \to \mathcal{O}_{T'}$ and $g_a : \mathcal{A}_T \to \mathcal{A}_{T'}$, we construct a transformed trajectory  
\[
\tau^g = \{(g_o \cdot o_t,\, g_a \cdot a_t,\, g_o \cdot o_{t+1},\, r_t)\}_{t=0}^N ,
\]  
which represents simulated experience in the symmetric augmented task $T'$.
The rewards remain unchanged due to their invariance under $g$.  

These augmented trajectories are integrated into policy optimization using Proximal Policy Optimization (PPO)~\cite{schulman2017ppo}.
Because both transition dynamics and rewards are $\mathbb{G}$-invariant, we can reuse state-transition likelihoods and advantage estimates from the original trajectories $\tau$ when updating on $\tau^g$~\cite{Mittal2024SymmetryCF}.
For each gradient step, a mini-batch consists of both original and transformed trajectories:  
\[
\mathcal{D}_{b} = \{\tau_i\}_{i=1}^{N_{b}} \cup \bigcup_{g \in \mathbb{G}} \{\tau_i^g\}_{i=1}^{N_{b}}.
\]  
This ensures balanced learning across direct and augmented tasks.  

The augmentation strategy applies broadly to robotic systems that exhibit morphological symmetries, which induce structured invariances in their kinematics, dynamics, and sensor measurements.
For instance, quadruped robots are approximately symmetric under left-right and front-hind reflections as well as a $180^\circ$ rotation ($\mathbb{K}_4$ subgroup), while humanoid robots typically exhibit left-right reflection symmetry ($\mathbb{C}_2$ subgroup)~\cite{ordonez2024morphological}.
With the same example, a quadruped trajectory that tracks a goal on the left with a failed left-front knee joint has a symmetric counterpart that tracks a mirrored goal on the right with a failure in the right-front knee.
By leveraging such symmetries, we can generate diverse training experiences offline, significantly improving training efficiency without additional environment interaction.

\subsection{Symmetry-Guided Memory Augmentation}

\begin{figure}
\centering
\includegraphics[width=1.0\linewidth]{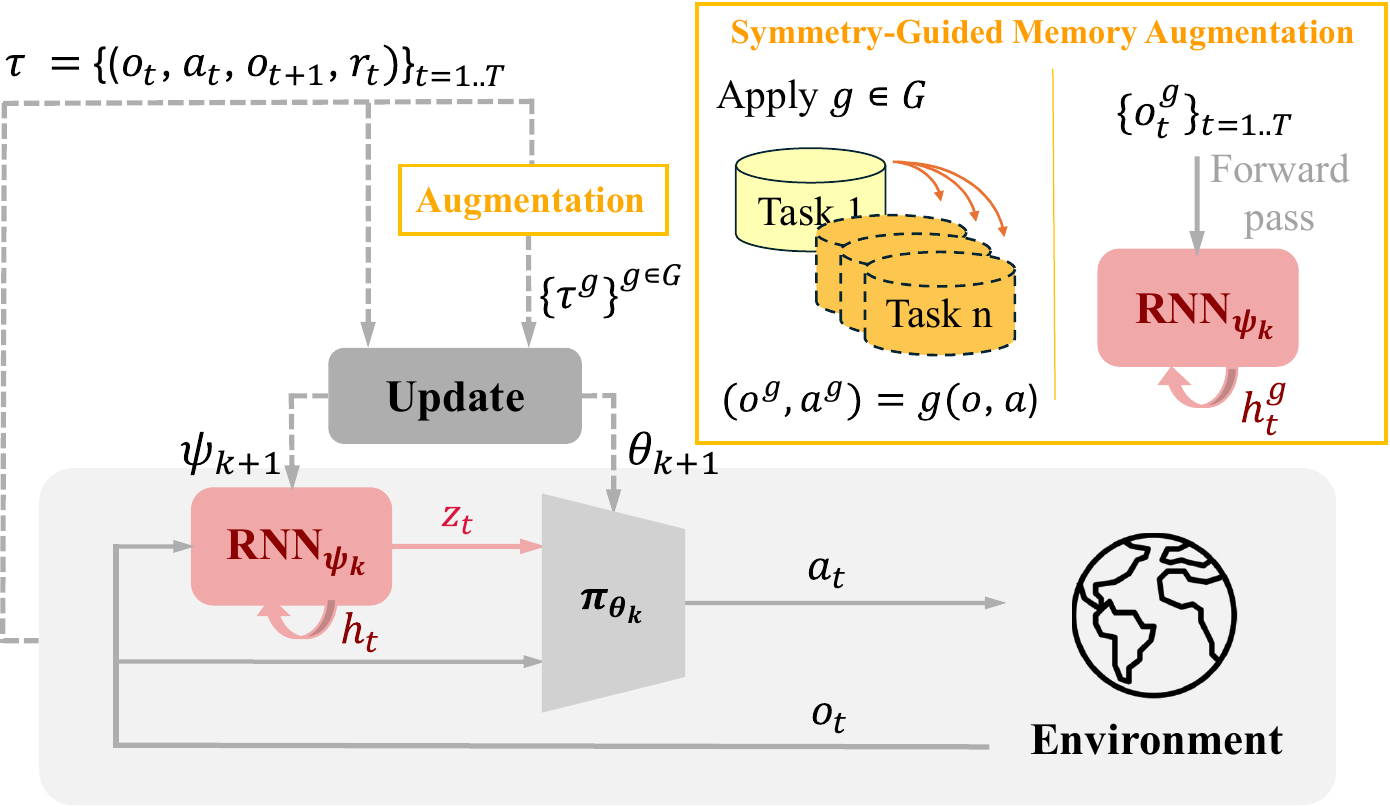}
\caption{Overview of our training framework. The policy is conditioned on latent embeddings $z_t$ inferred by an RNN that encodes task context from interaction history.}
\label{fig:training_framework}
\end{figure}

While experience augmentation improves generalization by exposing the agent to a richer variety of task conditions, the resulting variations, combined with the partial observability of the environment, may lead the agent to adopt overly conservative strategies as it struggles to infer the underlying task context.
To address this limitation, we extend augmentation to the agent’s memory by incorporating a recurrent neural network (RNN) into the training framework, as illustrated in~\figref{fig:training_framework}.
The RNN acts as an implicit task encoder: at each timestep, it updates its hidden state $h_t$ based on the observed trajectory, progressively capturing information about the latent environment context.
From this hidden state, a latent embedding $z_t$ is produced to condition the policy, enabling context-aware decision making.  

\begin{figure*}
\centering
\includegraphics[width=0.9\linewidth]{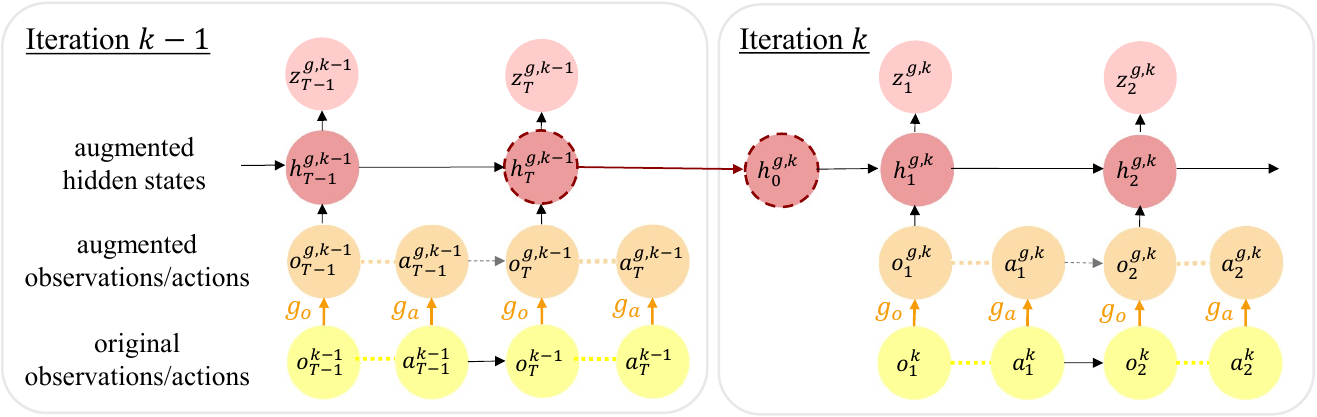}
\caption{Symmetry-Guided Memory Augmentation. A transformation $g = (g_o, g_a) \in \mathbb{G}$ is applied to observations $o_t$ and actions $a_t$ to generate augmented inputs $(o_t^g, a_t^g)$. The sequence of augmented observations is passed through the RNN, producing hidden states $(h_t^g)$ that encode task context for the augmented trajectory. At each policy update $k$, the initial hidden state $h_0^{g,k}$ is initialized from the final hidden state $h_T^{g,k-1}$ of the previous update, ensuring continuity and stable context retention.}
\label{fig:mem_aug}
\end{figure*}

In addition to augmenting trajectories, we augment the RNN’s hidden states to provide consistent memory for the augmented tasks.
As shown in~\figref{fig:mem_aug}, this is achieved by performing a forward pass with the transformed observation sequence $(o_t^g)_{t=0}^T$ through the RNN.
The resulting hidden states $(h_t^g)_{t=0}^T$ encode the evolving dynamics of the augmented trajectories.
To ensure continuity across updates, the initial hidden state $h_0^{g,k}$ for the $k$-th policy update is initialized from the final hidden state $h_T^{g,k-1}$ of the previous update.
This procedure ensures that the RNN’s hidden states consistently capture the cumulative task dynamics throughout training.  

The latent embeddings $(z_t^g)_{t=0}^T$ derived from the augmented trajectories allow the policy to infer unobservable context for both direct and augmented tasks.
In contrast, a feedforward policy without memory cannot recover such latent information from partial observations, often resulting in degraded performance when trained with augmented data.
By explicitly coupling symmetry-based augmentation with memory, our framework preserves task-specific adaptability while avoiding the pitfalls of context-unaware augmentation.
We next demonstrate that this combination enables efficient and robust learning across both simulation and hardware.





\section{EXPERIMENTS}
\label{sec:experiments}


\subsection{Environment and Task Setup}
We evaluate SGMA across eight locomotion experiments on the ANYmal D quadruped and the Unitree G1 humanoid robots using Isaac Lab~\cite{mittal2023orbit}.
The experiments include \textit{position tracking} and \textit{velocity tracking} tasks with variations introduced through joint failures and shifted payloads.
For each robot, training scenarios are categorized into \textbf{direct tasks}, which are explicitly encountered during training, and \textbf{augmented tasks}, which are generated through symmetry-guided augmentation.
The setup is summarized in Table~\ref{tab:task_setup}.

\begin{table}[t]
\centering
\caption{Locomotion experiment setup for SGMA. Experiments are conducted on \textit{position-tracking} and \textit{velocity-tracking} tasks using the ANYmal D quadruped and Unitree G1 humanoid robots. \textbf{Direct tasks} are scenarios explicitly encountered during training, while \textbf{augmented tasks} are simulated through symmetry-guided augmentation.}
\label{tab:task_setup}
\begin{tabular}{p{0.8cm}p{0.8cm}p{2.6cm}p{2.6cm}}
\toprule
\textbf{Robot} & \textbf{Variation} & \textbf{Direct Tasks} & \textbf{Augmented Tasks} \\
\midrule
ANYmal D & Joint failure & Failure of a left front leg joint (HAA, HFE, or KFE) & Failures in other leg joints \\
\cmidrule(lr){2-4}
         & Payload     & 30--40\,kg payload on left front base section & Same payload range on other base sections \\
\midrule
Unitree G1 & Joint failure & Failure of a left hip joint (pitch, roll, or yaw) & Failures in right hip joints \\
\cmidrule(lr){2-4}
           & Payload     & 40--50\,kg payload on left shoulder & Same payload range on right shoulder \\
           
\bottomrule
\end{tabular}
\end{table}
We evaluate our approach against two baselines: (i) \textbf{Rand} policies trained with all task variations (both direct and augmented tasks) exposed during training, and (ii) \textbf{NoAug} policies trained only on direct tasks without augmentation. We test both memory-enabled and feedforward (MLP) architectures, resulting in six variants: \textbf{SGMA}, \textbf{SGA-MLP} (symmetry-guided augmentation using MLP without memory~\cite{Mittal2024SymmetryCF, Su2024LeveragingSym}), \textbf{Rand-Memory}, \textbf{Rand-MLP}, \textbf{NoAug-Memory} and \textbf{NoAug-MLP}. Each approach is trained over five random seeds.

\subsection{Task Performance and Data Efficiency}
\begin{figure*}[ht]
    \centering
    \begin{subfigure}[b]{0.49\textwidth}
        \centering
        \includegraphics[width=\textwidth]{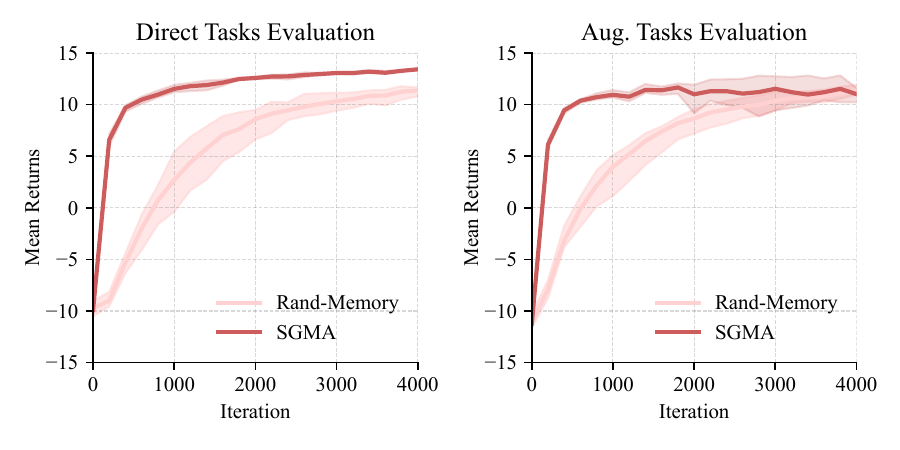}
        \caption{ANYmal position tracking under joint failure}
        \label{fig:anymal_agma_rand}
    \end{subfigure}
    \begin{subfigure}[b]{0.49\textwidth}
        \centering
        \includegraphics[width=\textwidth]{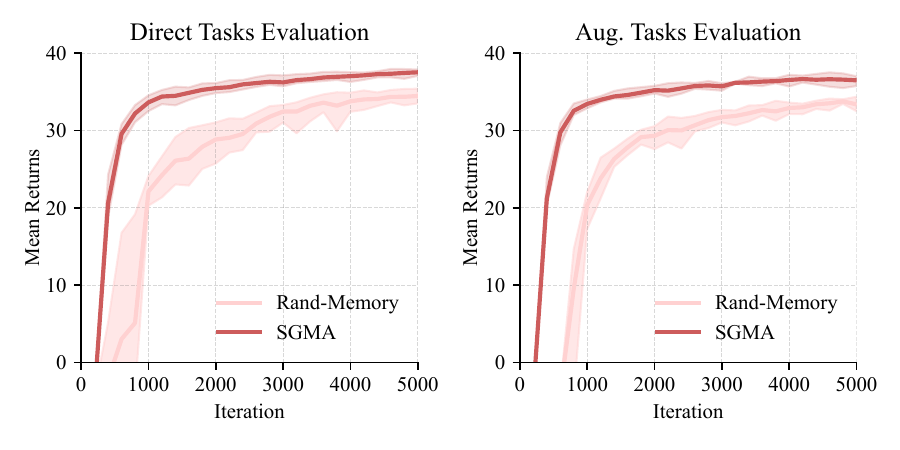}
        \caption{G1 velocity tracking under joint failure}
        \label{fig:g1_sgma_rand}
    \end{subfigure}
    \caption{Normalized mean episodic returns of \textbf{SGMA} and \textbf{Rand-Memory} policies on direct and augmented tasks, with error bars indicating minimum and maximum values.}
    \label{fig:sgm_rand}
\end{figure*}

\figref{fig:sgm_rand} presents a comparison between \textbf{SGMA} and \textbf{Rand-Memory} policies in two representative experiments.
In both cases, \textbf{SGMA} demonstrates markedly faster convergence, achieving stable performance substantially earlier than \textbf{Rand-Memory}, which requires considerably more training iterations to achieve comparable returns.
These results indicate that symmetry-guided augmentation accelerates learning by eliminating redundant interactions that randomization would otherwise expend on equivalent task variations.

\begin{figure}
\centering
\includegraphics[width=1.0\linewidth]{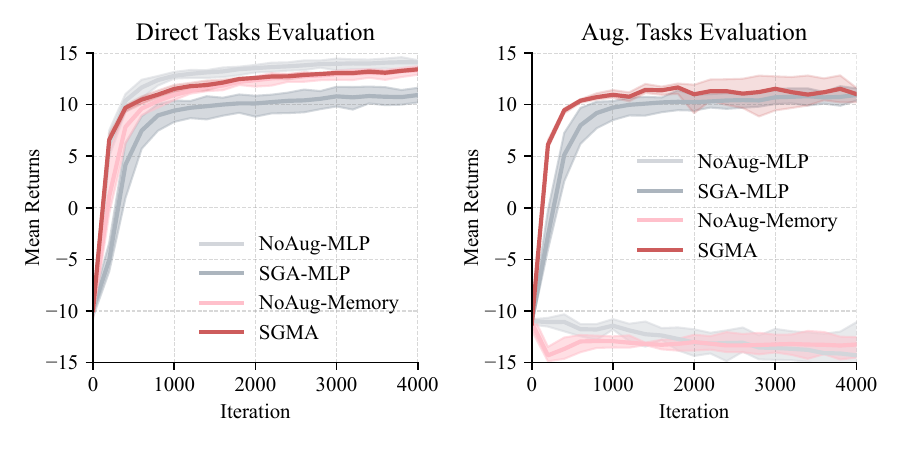}
\caption{Normalized mean episodic returns of policies trained \textbf{with} and \textbf{without augmentation} in quadruped position tracking under joint failure. Error bars indicate minimum and maximum values.}
\label{fig:sgma_id_plot}
\end{figure}

Next, we assess the role of the memory module in our framework.
\figref{fig:sgma_id_plot} compares policies trained with and without augmentation using both MLP-based and memory-enabled architectures in quadruped position tracking under joint failure experiment.
We observe a key trade-off with the MLP architecture.
Despite improving generalization on augmented tasks (\figref{fig:sgma_id_plot}, right), applying augmentation leads to a performance drop on direct tasks, with \textbf{SGA-MLP} performing worse than \textbf{NoAug-MLP} (\figref{fig:sgma_id_plot}, left). 
This indicates that naively applying augmentation to a feedforward policy in partially observable environments can compromise performance even on tasks explicitly seen during training, as the lack of context modeling forces the policy toward overly conservative strategies.
In contrast, \textbf{SGMA} effectively generalizes to augmented tasks while maintaining performance comparable to \textbf{NoAug-Rand} on direct tasks.
This suggests that the memory module effectively captures task context from partial observations and leverages it during experience augmentation, thereby preserving robust performance on seen tasks.

\begin{figure*}
\centering
\includegraphics[width=1.0\linewidth]{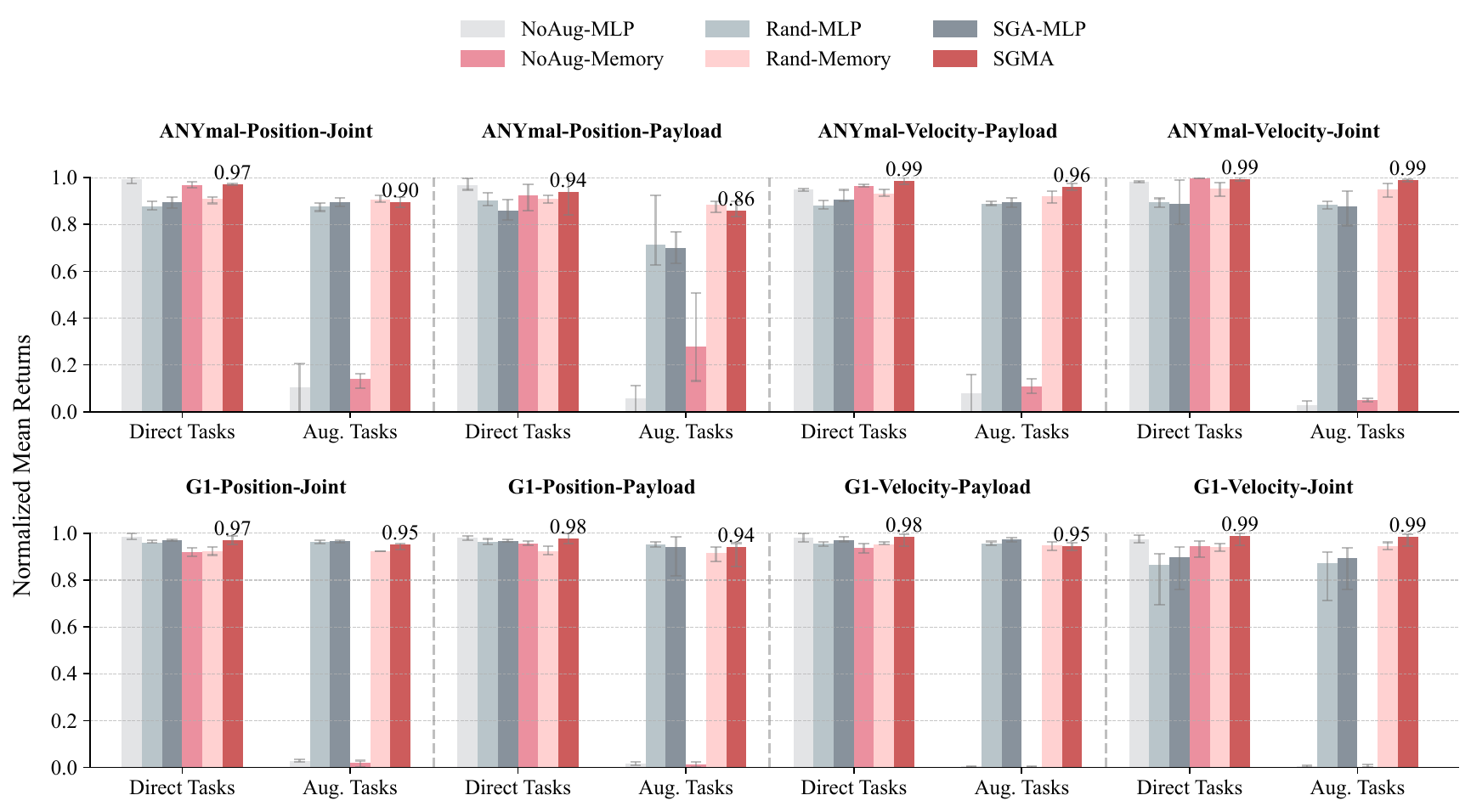}
\caption{Normalized mean episodic returns across all direct and augmented tasks. Error bars indicate minimum and maximum values.}
\label{fig:all_exp_results}
\end{figure*}

We extend this analysis to all experiments, with results summarized in \figref{fig:all_exp_results}.
Performance on direct tasks is consistently degraded for \textbf{SGA-MLP} and \textbf{Rand-MLP} compared to \textbf{NoAug-MLP}, particularly in quadruped experiments where task diversity is more pronounced.
This degradation again suggests that MLP policies, lacking the necessary task context, tend to adopt suboptimal strategies in partially observable settings when faced with increased variations introduced by augmentation or randomization.
In contrast, \textbf{SGMA} and \textbf{Rand-Memory} consistently match the performance of \textbf{NoAug-Memory} on direct tasks across all experiments.
By inferring latent task context from past experience, the memory module enables robust performance on seen tasks, even when trained under substantial task variations.
Critically, \textbf{SGMA} achieves performance comparable to \textbf{Rand-Memory} on both direct and augmented tasks. By generating variations offline through augmentation, our method delivers the same robust generalization as randomization, without requiring additional environment interactions.

\begin{figure*}
\centering
\includegraphics[width=1\linewidth]{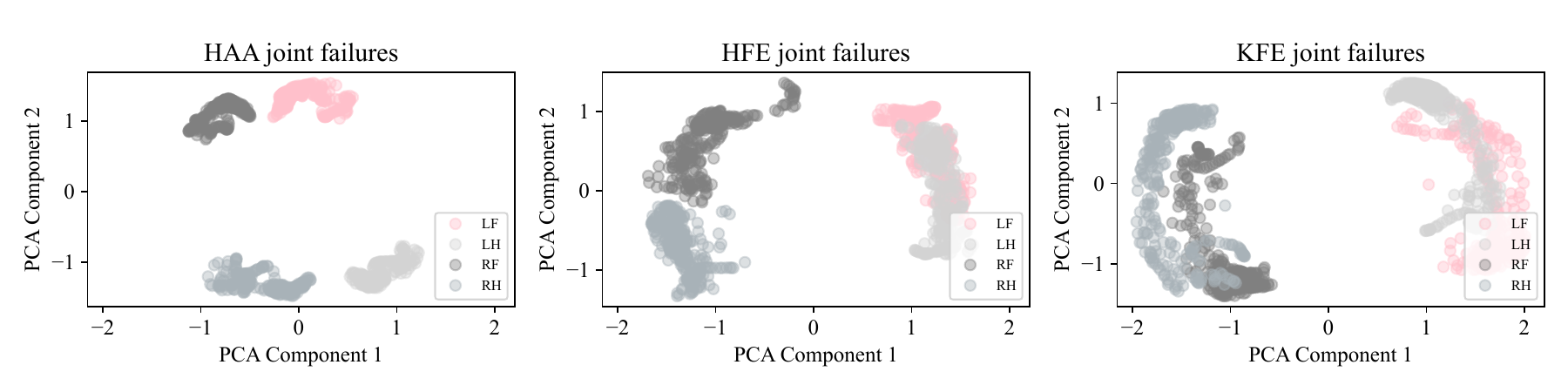}
\caption{PCA visualization of latent task embeddings $z$ (LF, in red) and $z_g$ (RF, LH, RH, in blue) for joint failures across HAA, HFE, and KFE joints. Distinct clustering indicates that the learned latent space effectively captures task-specific structure.}
\label{fig:latent_analysis}
\end{figure*}

To better understand how \textbf{SGMA} policy captures task-specific information, we analyze the latent task embeddings $z$ and $z_g$ generated by the memory module in the quadruped position tracking under the joint failures experiment.
\figref{fig:latent_analysis} presents a PCA visualization of the latent embeddings collected across all joint failure scenarios.
Notably, the embeddings reveal an emergent symmetry across different legs. This spatial symmetry in the latent representation aligns closely with the physical symmetry of the robot’s morphology and the underlying task structure.
This demonstrates the capacity of the memory module to infer and encode unobservable task context, which is crucial for the policy to adapt effectively across both directly seen and augmented tasks.

\subsection{Behavior Analysis with Memory Augmentation}

To better understand how the advantages of memory translate into concrete behavioral differences, we analyze locomotion strategies across different policies in quadruped position tracking under joint failures.

\begin{figure}
    \centering
    \includegraphics[width=\linewidth]{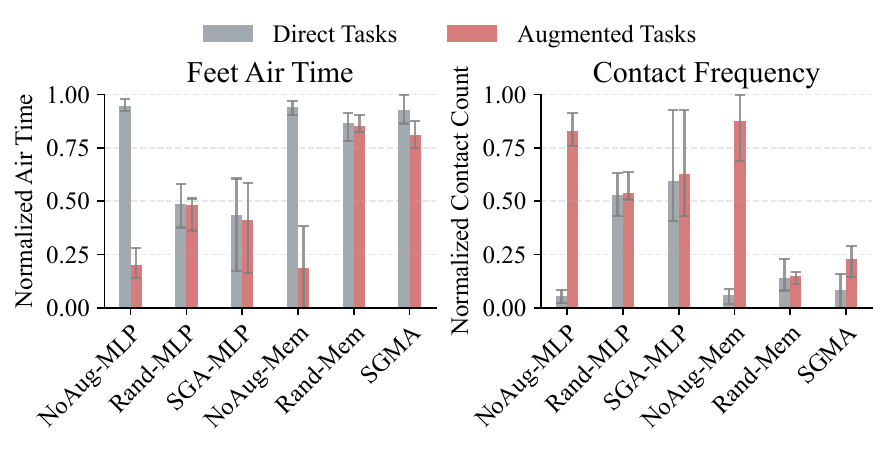}
    \caption{Motion analysis. Normalized episodic foot air time and contact frequency for different policies in quadruped position tracking under directly seen and augmented joint failures. Error bars indicate minimum and maximum values.}
    \label{fig:quantitative_motion_analysis}
\end{figure}

We first analyze foot motion statistics.
As shown in \figref{fig:quantitative_motion_analysis}, both \textbf{Rand-MLP} and \textbf{SGA-MLP} exhibit shorter air times and higher contact frequencies compared to \textbf{NoAug-MLP} on direct tasks.
This reflects a conservative locomotion strategy with smaller steps and frequent ground contacts.
The same pattern persists under augmented joint failures, suggesting that feedforward policies lack task-specific adaptation and default to overly cautious behaviors.
In contrast, memory-enabled policies—\textbf{SGMA} and \textbf{Rand-Memory}—achieve air time and contact frequency similar to \textbf{NoAug-Memory} on direct tasks, while maintaining comparable levels in augmented scenarios.
This robustness highlights the memory module’s ability to infer latent task context and support adaptive locomotion.  

\begin{figure*}
\centering
\includegraphics[width=\linewidth]{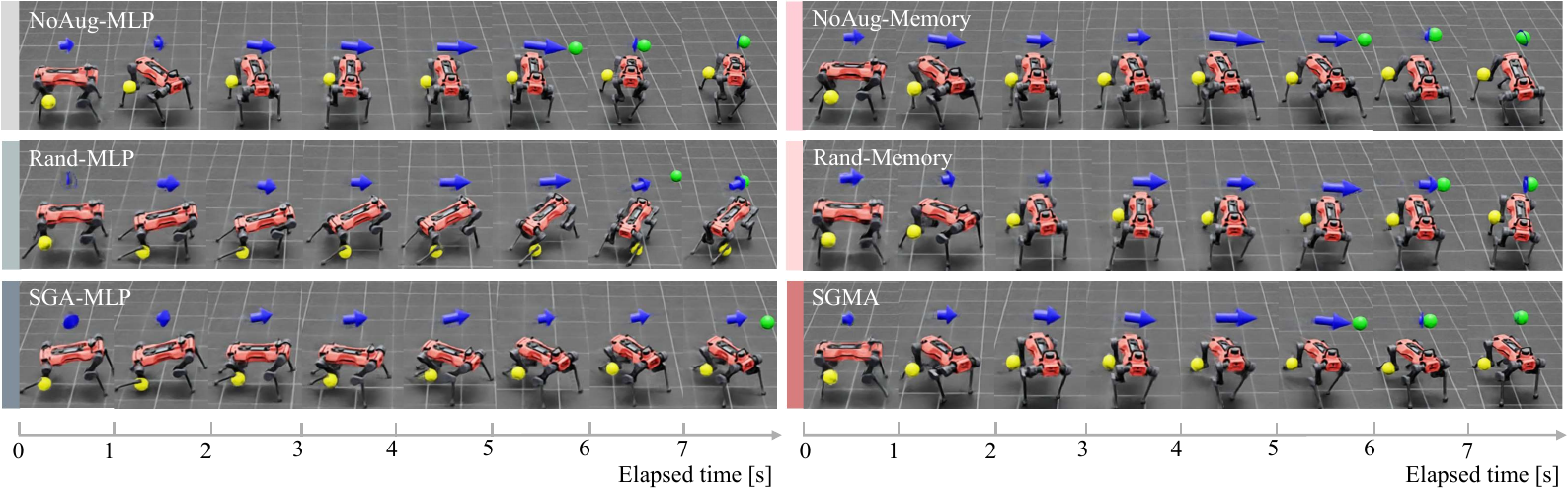}
\caption{Behavior analysis of different policies for quadruped position tracking under a left-front KFE joint failure (yellow marker). The blue arrow indicates base velocity; the green marker shows the goal. 
\textbf{\textit{Left:}} \textbf{NoAug-MLP} reorients sideways toward the goal and moves laterally, dragging the impaired leg for stability. In contrast, \textbf{Rand-MLP} and \textbf{SGA-MLP} walk cautiously forward with small steps, often stumbling and drifting. 
\textbf{\textit{Right:}} Memory-enabled agents (\textbf{SGMA} and \textbf{Rand-Memory}) actively adjust their body orientation, compensating for the impaired joint and maintaining stable locomotion.}
\label{fig:behavior_analysis}
\end{figure*}

We also qualitatively examine goal-tracking behaviors. \figref{fig:behavior_analysis} illustrates representative trajectories under a left-front KFE joint failure, a scenario directly encountered during training.
Agents trained with \textbf{Rand-MLP} and \textbf{SGA-MLP} walk cautiously toward the goal using small steps, often stumbling or drifting in orientation.
This behavior aligns with the quantitative findings of shorter air time and higher contact frequency.
These agents appear to reduce overall joint usage uniformly rather than compensating specifically for the failed joint.
By contrast, \textbf{NoAug-MLP} and \textbf{NoAug-Memory} adopt more targeted strategies: they reorient the body sideways toward the goal and then move laterally, dragging the impaired leg to reduce reliance on it (upper row of \figref{fig:behavior_analysis}).
This adjustment yields more stable and effective tracking.
Notably, \textbf{SGMA} and \textbf{Rand-Memory} display similar adaptive behaviors, actively compensating for the impaired joint.
This again demonstrates the critical role of memory in inferring task context and enabling targeted adaptation.

\subsection{Hardware Experiments}

\begin{figure}[!t]
    \centering
    \begin{subfigure}[t]{0.95\columnwidth}
        \centering
        \includegraphics[width=\linewidth]{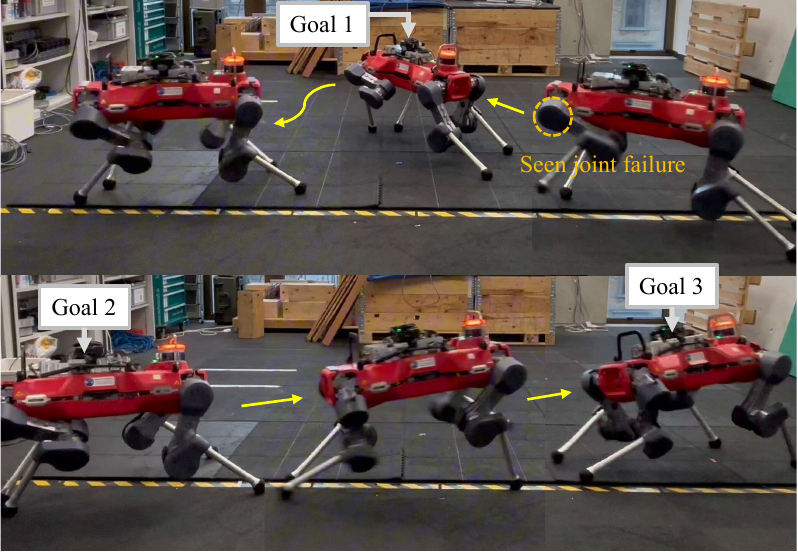}
        \caption{\textbf{SGMA}-trained robot successfully tracks multiple goals under a seen joint failure.}
        \label{fig:hardware_id_mem_aug}
    \end{subfigure}
    \vspace{0.5em}
    \begin{subfigure}[t]{0.95\columnwidth}
        \centering
        \includegraphics[width=\linewidth]{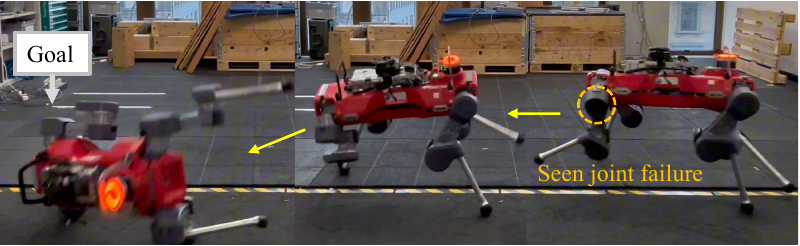}
        \caption{\textbf{SGA-MLP}-trained robot shows limited adaptability to the seen joint failure and loses balance while tracking goals.}
        \label{fig:hardware_id_base_aug}
    \end{subfigure}
    \caption{Hardware experiments on ANYmal D. Quadruped position tracking performance under an LF HAA joint failure, a scenario explicitly included during training.}
    \label{fig:hardware_results}
\end{figure}

We deploy our \textbf{SGMA} policy on the ANYmal D robot for position tracking under joint failure.
To facilitate sim-to-real transfer, we introduce randomization in terrain profiles with a noise range of \([0.0, 0.05]\) during training.
Despite the inherent asymmetries and deviations of real-world environments from simulation assumptions, our policy demonstrated robust, zero-shot generalization.
The robot successfully tracked multiple goals under a joint failure (RH HAA) that was never explicitly encountered during training but was simulated solely through symmetry-guided augmentation (see~\figref{fig:intro}).
The policy also maintains robust performance under a directly trained-on joint failure (LF HAA), as shown in~\figref{fig:hardware_id_mem_aug}.
Additionally, we also test \textbf{SGA-MLP} policy trained under the same terrain profiles. 
As shown in~\figref{fig:hardware_id_base_aug}, the robot exhibited limited adaptability under ID joint failure, often losing balance during goal tracking.
These results confirm that SGMA not only improves training efficiency in simulation but also transfers effectively to hardware, enabling robust adaptation even under previously unseen joint failures.
Please visit our webpage for the experiment videos.

\section{LIMITATIONS}
The main limitation of our work lies in the reliance on prior knowledge of the task structure, which is essential for augmenting the training experience to simulate diverse task conditions. 
While leveraging symmetry enables efficient generation of diverse training experiences without additional environment interactions, it constrains the method to domains where such symmetry assumptions are valid and well-understood. 
That said, incorporating domain knowledge offers a principled way to inject useful inductive bias, allowing for more targeted and sample-efficient training.
In the future, we aim to broaden the applicability of our method by exploring more flexible augmentation techniques to simulate a wider range of task conditions and further enhance training efficiency in less structured environments.

\section{CONCLUSION}
In this work, we present SGMA, a principled method that enhances policy training efficiency for legged locomotion by combining structured experience augmentation with memory-based context inference. 
Our approach leverages the inherent symmetries of robots and tasks to generate additional, physically consistent training experiences without the need for new environment interactions. This directly addresses the high interaction costs and sample inefficiency of traditional randomization-based methods, which rely on collecting costly data for every possible scenario.
Crucially, the memory module proves essential in our framework, allowing the agent to infer task context from past interactions in partially observable settings with substantial task diversity. 
This facilitates adaptation across tasks and overcomes the common pitfalls of context-unaware policies, which tend to adopt over-conservative strategies in highly randomized environments.

Through extensive simulation experiments, we demonstrate that our approach achieves robust generalization to augmented tasks while maintaining strong performance on explicitly trained ones. Notably, SGMA matches the performance of randomization-based methods on both seen and augmented tasks with enhanced efficiency, without requiring explicit environment interactions for the latter. 
Furthermore, we validate the sim-to-real transfer of our policy on a real quadruped robot, demonstrating successful adaptation to both seen and augmented tasks in real-world scenarios.
We believe this work provides a practical and data-efficient route toward training adaptive policies, inspiring further research in efficient policy learning for advanced robotics tasks.




\section*{APPENDIX}
\Tabref{tab:training_details} summarizes the hyperparameter settings used in our experiments.
We refer to our webpage for more implementation details.

\begin{table}[ht]
\centering
\caption{Hyperparameters of our experiments}
\label{tab:training_details}
\begin{tabular}{
    >{\raggedright\arraybackslash}p{2.5cm}
    >{\centering\arraybackslash}p{1.2cm}
    >{\centering\arraybackslash}p{0.8cm}
    >{\centering\arraybackslash}p{1.2cm}
    >{\centering\arraybackslash}p{0.8cm}
}
\toprule
\textbf{Hyperparameter} & \multicolumn{2}{c}{\textbf{Position Tracking}} & \multicolumn{2}{c}{\textbf{Velocity Tracking}} \\
\cmidrule(lr){2-3} \cmidrule(lr){4-5}
 & \textbf{ANYmal} & \textbf{G1} & \textbf{ANYmal} & \textbf{G1} \\
\midrule
\multicolumn{5}{l}{\textbf{PPO hyperparameters}} \\
Parallel environments & 4096 & 4096 & 4096 & 4096 \\
Steps per environment & 48 & 48 & 24 & 24 \\
Epochs per update & 5 & 5 & 5 & 5 \\
Minibatches per epoch & 4 & 4 & 4 & 4 \\
Discount rate ($\gamma$) & 0.99 & 0.99 & 0.99 & 0.99 \\
GAE parameter & 0.95 & 0.95 & 0.95 & 0.95 \\
Value loss coefficient & 1.0 & 1.0 & 1.0 & 1.0 \\
Clipping parameter & 0.20 & 0.20 & 0.20 & 0.20 \\
Entropy coefficient & 0.005 & 0.003 & 0.005 & 0.008 \\
Max. gradient norm & 1.0 & 1.0 & 1.0 & 1.0 \\
Initial learning rate & 0.001 & 0.001 & 0.001 & 0.001 \\
Target KL & 0.01 & 0.01 & 0.01 & 0.01 \\
Optimizer & Adam & Adam & Adam & Adam \\
\midrule
\multicolumn{5}{l}{\textbf{Policy network architectures}} \\
Policy MLP hidden layers & 128,128,\allowbreak128 & 256,128,\allowbreak128 & 128,128,\allowbreak128 & 256,128,\allowbreak128 \\
Critic MLP hidden layers & 128,128,\allowbreak128 & 256,128,\allowbreak128 & 128,128,\allowbreak128 & 256,128,\allowbreak128 \\

MLP activations & ELU & ELU & ELU & ELU \\
\midrule
\multicolumn{5}{l}{\textbf{Memory module}} \\
RNN type & LSTM & LSTM & LSTM & LSTM \\
RNN hidden size & 256 & 256 & 256 & 512 \\
RNN hidden layers & 1 & 1 & 1 & 1 \\
\bottomrule
\end{tabular}
\end{table}




\bibliographystyle{IEEEtran}
\bibliography{IEEEabrv,main}

\end{document}